\documentclass[10pt,twocolumn,letterpaper]{article}

\usepackage{cvpr}
\usepackage{times}
\usepackage{epsfig}
\usepackage{graphicx}
\usepackage{amsmath}
\usepackage{amssymb}

\usepackage{color}
\usepackage{graphicx}
\usepackage{ulem}
\usepackage{ifthen}
\usepackage{array,multirow}
\usepackage{adjustbox}

\definecolor{gray}{rgb}{0.5,0.5,0.5} 
\definecolor{green}{rgb}{0, 0.4, 0} 
\definecolor{orange}{rgb}{1, 0.5, 0} 	
\definecolor{mahogany}{rgb}{0.75, 0.25, 0.0}
\definecolor{purple}{rgb}{0.6, 0, 0.6}
\definecolor{purple}{rgb}{0.6, 0, 0.6}
\definecolor{darkgreen}{rgb}{0, 0.4, 0.4} 
\definecolor{frenchblue}{rgb}{0.0, 0.45, 0.73}
\definecolor{what_color}{rgb}{0.7, 0.4, 0.3}

\newboolean{revising}
\setboolean{revising}{false}
\ifthenelse{\boolean{revising}}
{
	\newcommand{\ignore}[1]{}
    \newcommand{\stan}[1]{\textcolor{green}{#1}}
	\newcommand{\stanreplace}[2]
    {\textcolor{green}{#2}}
	\newcommand{\drliu}[1]{\textcolor{frenchblue}{#1}}
	\newcommand{\drliureplace}[2]{\textcolor{frenchblue}{#2}}
	\newcommand{\yenchen}[1]{\textcolor{darkgreen}{#1}}
	
	\newcommand{\hu}[1]{\textcolor{orange}{#1}}
	\newcommand{\hureplace}[2]{\textcolor{orange}{#2}}
	\newcommand{\sunmin}[1]{\textcolor{purple}{#1}}

	\newcommand{\chengreplace}[2]{\textcolor{what_color}{#2}}
} {
	\newcommand{\ignore}[1]{}
    \newcommand{\stan}[1]{#1}
	\newcommand{\stanreplace}[2]{#2}
	\newcommand{\drliu}[1]{#1}
	\newcommand{\drliureplace}[2]{#2}
	\newcommand{\yenchen}[1]{#1}
	
	\newcommand{\hu}[1]{#1}
	\newcommand{\hureplace}[2]{#2}
	\newcommand{\sunmin}[1]{#1}

	\newcommand{\chengreplace}[2]{#2}
}

\newcommand{\de}{$^\circ$ }

\newcommand{\cutsectionup}{\vspace*{-0.1in}}
\newcommand{\cutsectiondown}{\vspace*{-0.1in}}

\newcommand{\cutsubsectionup}{\vspace*{-0.11in}} 
\newcommand{\cutsubsectiondown}{\vspace*{-0.08in}} 

\newcommand{\cuttableup}{\vspace*{-0.2in}}

\usepackage[pagebackref=true,breaklinks=true,letterpaper=true,colorlinks,bookmarks=false]{hyperref}

\newcommand*{\affaddr}[1]{\normalsize#1}
\newcommand*{\affmark}[1][*]{\normalsize\textsuperscript{#1}}
\newcommand*{\email}[1]{\footnotesize\texttt{#1}}

\cvprfinalcopy 

\def\cvprPaperID{1277} 

\ifcvprfinal\fi
\begin{document}

\title{Deep 360 Pilot: Learning a Deep Agent for Piloting through 360$^{\circ}$ Sports Videos}

\author{
\normalsize
Hou-Ning Hu\affmark[1]\thanks{indicates equal contribution}  ,
Yen-Chen Lin\affmark[1]\footnotemark[1]  ,
Ming-Yu Liu\affmark[2],  
Hsien-Tzu Cheng\affmark[1],
Yung-Ju Chang\affmark[3],  
Min Sun\affmark[1] \\
\affaddr{\affmark[1]National Tsing Hua University} \hspace{1.5mm} 
\affaddr{\affmark[2]NVIDIA research} \hspace{1.5mm} 
\affaddr{\affmark[3]National Chiao Tung University} \\
\email{\{eborboihuc, hsientzucheng\}@gapp.nthu.edu.tw} \hspace{1.5mm} \email{armuro@cs.nctu.edu.tw},\\
\email{\{yenchenlin1994, sean.mingyu.liu\}@gmail.com} \hspace{1.5mm} \email{sunmin@ee.nthu.edu.tw}
}
\maketitle
\thispagestyle{empty}

\vspace{-6mm}

\begin{abstract}
Watching a 360$^{\circ}$ sports video requires \drliureplace{a user to continuously select views -- viewing angles with a Nature-Field-of-View (NFoV) (see cyan box in Fig.~\ref{fig.teaser}).}{a viewer to continuously select a viewing angle, either through a sequence of mouse clicks or head movements.}
To relieve the \drliureplace{user}{viewer} from this ``360 piloting" task, we propose ``deep 360 pilot'' -- a deep learning-based agent for piloting through 360$^{\circ}$ sports videos automatically. At each frame, the agent observes a panoramic image and has the knowledge of previously selected \drliureplace{views}{viewing angles}. The task of the agent is to shift the current viewing angle (i.e. action) to the next preferred \drliureplace{viewing angle}{one} (i.e., goal)\drliureplace{, while fixing the NFoV}{}.
We propose to directly learn \drliureplace{the}{an} online policy of the agent \drliu{from data}. Specifically, we leverage \drliureplace{the}{a} state-of-the-art object detector to \drliureplace{select}{propose} a few candidate objects of interest (yellow boxes in Fig.~\ref{fig.teaser}). Then, a \drliureplace{Recurrent Neural Network (RNN)}{recurrent neural network} is used to select the main object (green dash boxes in Fig.~\ref{fig.teaser}). Given the main object and previously selected \drliureplace{views}{viewing angles}, our method regresses a shift in viewing angle to move to the next \drliureplace{view}{one}. \sunmin{We use \drliureplace{standard and}{the} policy gradient \drliu{technique} to jointly train our pipeline\stanreplace{:}{,} by minimizing\stan{:} (1) a regression loss measuring the distance between the selected and ground truth viewing angles, \drliureplace{and (2)}{(2)} a smoothness loss \drliureplace{to encourage}{encouraging} smooth transition in viewing angle\drliureplace{; as well as by}{, and (3)} maximizing an expected reward \drliu{of focusing on a foreground object}}. To evaluate our method, we \stanreplace{collect}{built} a new 360-Sports video dataset consisting of five \drliu{sports} domains. We train\stan{ed} domain-specific agents and achieve\stan{d} the best performance on \drliu{viewing angle} selection accuracy and \hureplace{transition smoothness}{users' preference} \drliureplace{in viewing angle}{}compared to \cite{supano2vid} and other baselines.



\end{abstract}
\vspace{-3mm}

\cutsectionup
\section{Introduction}\label{sec.Intro}
\cutsectiondown

\drliureplace{360-degree (referred to as 360$^\circ$)}{360$^\circ$} video gives a viewer immersive experiences through displaying full surroundings of a camera in a spherical canvas, which differentiates itself from traditional multimedia. As consumer- and production-grade 360$^\circ$ cameras become readily available, 360$^\circ$ videos are captured every minute. Moreover, the promotion of 360$^\circ$ videos by social media giants including YouTube and Facebook further boosts their fast adoption. It is expected that 360$^\circ$ videos will become a major \drliureplace{multimedia for transmitting visual contents}{video format} in the near future. Studying how to display 360$^\circ$ videos to a human viewer, who has a limited field of visual attention, emerges as an increasingly important problem.

\begin{figure}[t!]
\begin{center}
\includegraphics[width=0.975\linewidth]{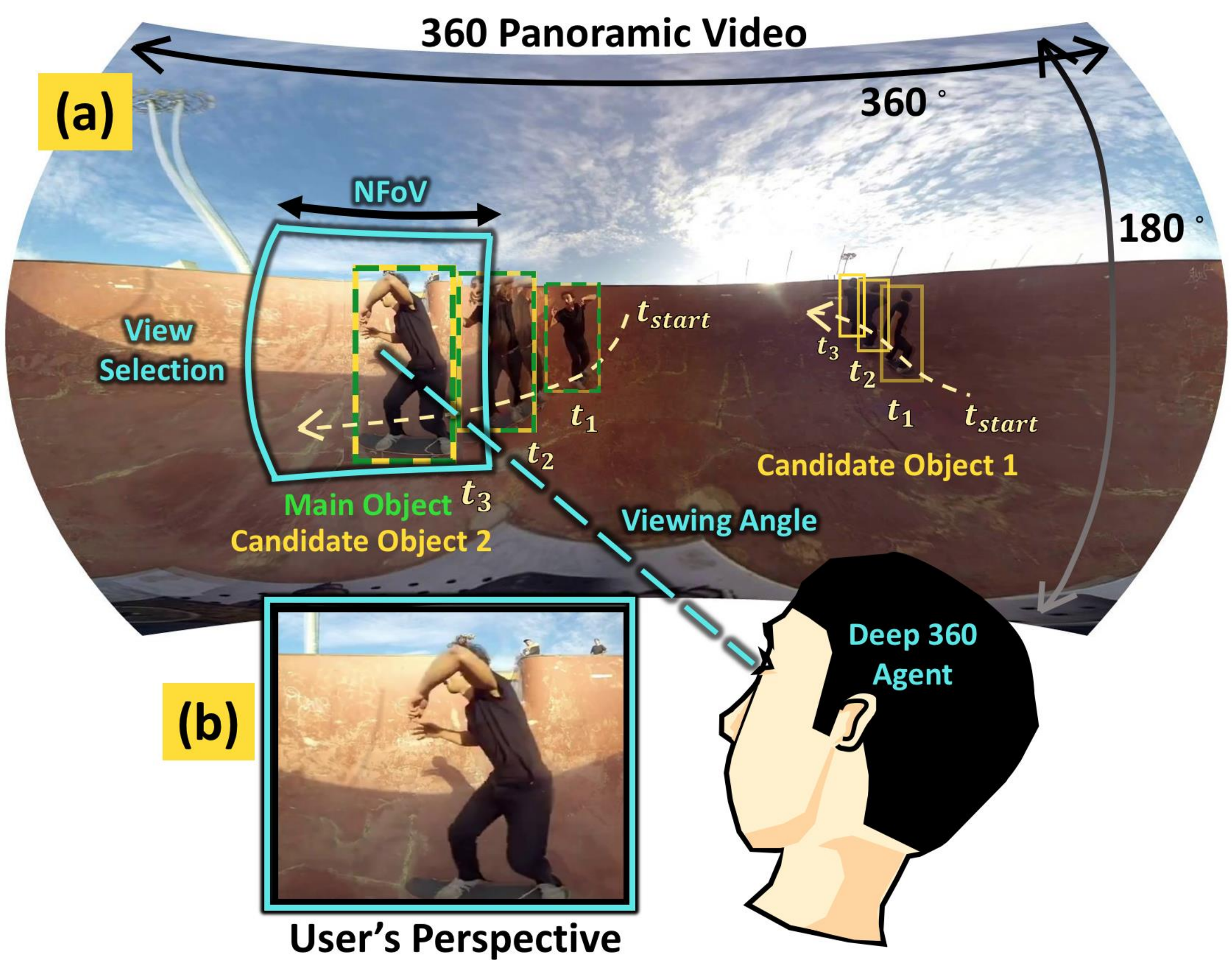}
\end{center}
\vspace{-3mm}
\caption{Panel (a) overlaps three panoramic frames sampled from a 360\de skateboarding video with two skateboarders. One skateboarder is more active than the other in this example. For each frame, the proposed ``deep 360 pilot" selects a view -- a viewing angle, where a Natural Field of View (NFoV) (cyan box) is centered at. It first extracts candidate objects (yellow boxes), and then selects a main object (green dash boxes) in order to determine a view (just like a human agent). Panel (b) shows the NFoV from a viewer's perspective.
}\label{fig.teaser}
\vspace{-4mm}
\end{figure}

Hand Manipulation (HM) and Virtual Reality (VR) are two main ways for displaying 360$^\circ$ videos on a device with a Natural Field of View (NFoV) (typically a 60\de to 110\de FoV as shown in Fig.~\ref{fig.teaser}). In HM, a viewer navigates a 360$^\circ$ video via a sequence of mouse clicks; whereas, in VR, a viewer uses embedded motion sensors in a VR headset for navigation. Note that both HM and VR require a viewer to select a viewing angle at each frame, while the FoV is defined by the device. \drliureplace{This is referred to as view selection.}{}
For sports videos, such a selection mechanism could be cumbersome because ``foreground objects" of interest change their locations continuously. In fact, a recent study~\cite{LinCHI17} showed that both HM and VR can cause a viewer to feel discomfort. Just imagine how hard it is to follow an X-game skateboarder in a 360$^\circ$ video. Hence, a mechanism to automatically navigate a 360$^\circ$ video in a way that captures most of the interest events for a viewer would be beneficial.

Conceptually, \drliureplace{when watching a 360$^\circ$ video, a}{a  360$^\circ$-video} viewer is \drliureplace{like an}{a human} agent: at each frame, the agent observes a panoramic image (i.e., the observed state) and steers the viewing angle (i.e. the action) to cover the next preferred \drliureplace{view}{viewing angle} (i.e., the goal). We refer to this process as \textit{360 piloting}. Based on this analogy and, more importantly, \hureplace{for the purpose of relieving}{to relieve} the viewer\drliureplace{'s pain of constantly manipulating a 360\de video}{~from constantly steering the viewing angle} while watching \drliureplace{the}{360\de} video\drliu{s}, we argue for an intelligent agent that can automatically piloting through 360\de sports videos.

\drliureplace{It is noteworthy that using}{Using} an automatic mechanism for displaying video contents is not a new idea. For example, \drliureplace{researchers have proposed }video summarization--condensing a long video into a short summary video~\cite{Truong:2007}\drliureplace{, which}{--}has been used in reviewing hourly long surveillance videos. 
However, while a video summarization algorithm makes binary decisions on whether to select a frame or not, an agent for 360 piloting needs to operate on a spatial space to steer the viewing angle to \stanreplace{cover}{consider} events of interest in a 360$^\circ$ video. On the other hand, in virtual cinematography, \sunmin{most camera manipulation  tasks are} performed within relatively simpler virtual environments~\cite{ChristiansonAHSWC96,He:1996,elson_lightweight_2007,Mindek:2015} and \hureplace{therefore}{} there is no need to deal with \stanreplace{difficult perception problem}{viewers' perception difficulty} because 3-D positions and poses of all entities are known.  
However, a practical agent for 360 piloting needs to directly work with raw 360$^\circ$ videos. For displaying 360$^\circ$ videos, Su et al.~\cite{supano2vid} proposed \stanreplace{first}{firstly} detecting candidate events of interest in the entire video, and then apply\stan{ing} dynamic programming to link \stanreplace{the}{detected} events. 
\stanreplace{As}{However, as} \stanreplace{requiring}{this method requires} observing an entire video, it is non-suited for video streaming applications such as foveated rendering~\cite{Patney16}. We argue that being able to make a selection based on \stan{the}{} current and previous frames (like a human agent does) is critical for 360 piloting.
Finally, both \cite{supano2vid} and recent virtual cinematography works~\cite{chen2016learning,chen2015mimicking}  aim for smooth \drliu{viewing angle transition. Such transition should also be enforced for 360$^\circ$ piloting.}  \drliureplace{ transition in viewing angle. Hence, the agent should also aim at steering viewing angle smoothly.}{}

We propose ``deep 360 pilot''---a deep learning-based agent that navigates a 360$^{\circ}$ sports video in a way that smoothly captures interesting moments in the video.
Our ``deep 360 pilot" agent not only follows foreground objects of interest but also steers the viewing angle smoothly \stanreplace{for viewers' comfort}{to increase viewers' comfort}. 
We propose the following online pipeline to learn an online policy from human agents to model how a human agent takes actions in watching sports videos. First, because in sports videos foreground objects are those of viewers' interest, we leverage a state-of-the-art object detector~\cite{ren2015faster} to identify candidate objects of interest. Then, a Recurrent Neural Network (RNN) is used to select the main object among candidate objects. Given the main object and previously selected \drliureplace{views}{viewing angles}, our method predict\stan{s} how to steer the viewing angle to the preferred \drliureplace{view}{one} by learning a regressor. 
\stanreplace{Except the object detector}{In addition}, our pipeline is jointly trained with the following functions: (1) a regression loss measuring the distance between the selected and ground truth viewing angles, \drliureplace{and} (2) a smoothness loss to encourage smooth transition in viewing angle, \drliureplace{as well as}{and (3)} an expected reward \drliu{of focusing on a foreground object}. \drliureplace{Note that we use both standard and}{We used the} policy gradient \drliu{technique}~\cite{Williams1992} to train the pipeline since \drliureplace{our pipeline}{it} involves making an intermediate discrete decision corresponding to selecting the main object.
To evaluate our method, we collect\stan{ed} a new 360$^{\circ}$ sport\stan{s} video dataset consisting of five domains and train\stan{ed} an agent for each domain (referred to as 360-Sports).
These domain-specific agents achieve the best performance in regression accuracy and transition smoothness in viewing angle.

Our main contributions \drliureplace{can be summarized}{are} as follows:
(1) \drliureplace{We are the first to develop an online agent acting like human agents in selecting views when watching 360$^\circ$ videos, a characteristic critical to model especially for watching 360$^\circ$ streaming videos and predicting views for foveated VR rendering.}{We develop the first human-like online agent for automatically navigating 360$^\circ$ videos for viewers. The online processing nature suits the agent for streaming videos and predicting views for foveated VR rendering.}
(2) \drliureplace{We propose an jointly trainable policy pipeline based on a deep neural network. Our policy network is trained using policy gradient to handle non-differentiable main object selection layer.}{We propose a jointly trainable pipeline for learning the agent. Since the main object selection objective is non-differentiable, we employ a policy gradient technique to optimize the pipeline.}
(3) \drliureplace{The objective of our agent considers both view selection accuracy and transition smoothness in viewing angle.}{Our agent considers both viewing angle selection accuracy and transition smoothness.}
(4) We \drliureplace{collected}{build} the first 360$^\circ$ sports videos dataset to train and evaluate our ``deep 360 pilot" agent.
\cutsectionup
\section{Related Work}\label{sec.RW}
\cutsectiondown

We \drliureplace{describe}{review} related works in video summarization, saliency detection, and virtual cinematography. 

\cutsubsectionup
\subsection{Video Summarization}
\cutsubsectiondown
\drliureplace{
There is a large \stan{body of} literature~\cite{Truong:2007} on video summarization\drliureplace{-- condensing videos into short summaries.}{.} Here we \stanreplace {mention the most relevant few works.}{selectively review several most relevant works\drliureplace{in our view.}{.}}
}
{We selectively review several most relevant video summarization works from a large body of literature~\cite{Truong:2007}.}

\noindent\textbf{Important frame sampling.}
\cite{LiuTPAMI,LiuCVPR00,ZhanCSVT05,CVPR13_Khosla} proposed to sample a few important frames as the summary of a video.
\cite{potapov2014category,sun2014ranking,Yao_2016_CVPR} focused on sampling domain-specific highlights.
\cite{BinLiveLight,sun2014ranking,NIPS2014_5413} proposed weakly-supervised methods to select important frames. Recently, a few \drliu{deep learning-based} methods~\cite{Zhang_2016_CVPR,Zhang_2016_CVPR,ZhangCSG16} \drliureplace{based on deep learning}{} have shown impressive performance.
\cite{WebSyn07,L2S06,MSunMontage} focused on extracting highlights and generating synopses which showed several spatially non-overlapping actions from different times of a video. Several methods \cite{StoryB06,Cliplets12} involving user interaction have also been proposed in the graphics and the HCI communities.

\noindent\textbf{Ego-centric video summarization.}
In ego-centric videos, cues from hands and objects become easier to extract compare to third-person videos.
\cite{EgoS12} proposed \stanreplace{to summarize a video according to}{video summarization based on the}  interestingness and diverseness of objects and faces.
\cite{EgoS13} further proposed  \stanreplace{to track objects and measure how influential each frame is.}{tracking objects and measuring the influence of individual frames.}
\cite{Kopf:FHV} proposed a novel approach to speed-up ego-centric videos while removing unpleasant camera movements.

\drliureplace{However, most video summarization methods are concerned whether to select a frame. View selection in 360$^\circ$ videos is very different because a video frame is panoramic and thus contains more candidate views potentially interesting to the viewer.}
{In contrary to most video summarization methods which concern whether to select a frame or not, a method for 360 piloting concerns which viewing angle to select for each panoramic frame in a 360$^\circ$ video.}

\cutsubsectionup
\subsection{Saliency Detection}
\cutsubsectiondown
Many methods have been proposed to detect salient regions typically measured by human gaze. 
\cite{liu2011learning,HarelKP06,AchantaHES09,perazzi2012saliency,wang2016learning,zhang2016exploiting,perazzi2012saliency} focused on detecting salient regions on images. Recently, 
\cite{Liu_2016_CVPR,Jetley_2016_CVPR,mlnet2016,pan2016shallow,Bruce_2016_CVPR,Wang_2016_CVPR,WangWLZR16,TangW16} \drliureplace{\stanreplace {based on}{leveraging}}{leveraged} deep learning \drliureplace{have achieved}{and achieved} significant performance gain.
For videos,
\cite{MSali09,STSali08,MahadevanTPAMI,SeoJOV,DIEM,lee2015low} relied on low-level
appearance and motion cues as inputs. In addition, 
\cite{Judd09,GofermanTPAMI,rudoy2013learning,MatheSminchisescuPAMI2015,GazePlus} included information about face, people, objects, or other contexts.
Note that saliency detection methods do not select views directly, but output a saliency score map. \hu{Our method is also different to visual attention methods for object detection~\cite{mps-rlcvpr16,ba-attention-2015,mnih-attention-2014} in that it considers view transition smoothness as selecting views, which is crucial for video watching experience.}

\noindent\textbf{Ranking foreground objects of interest.}
Since regions of interest in sports videos are typically foreground objects, \cite{MSunMontage} propose\stan{d} to use an object detector~\cite{BourdevMalikICCV09} to extract candidate objects of interest, then rank the saliency of these candidate objects. For 360 piloting, we propose a similar baseline which first detects objects using RCNN~\cite{ren2015faster}, then select the viewing angle focusing on the most salient object according to a saliency detector~\cite{zhang2016exploiting}. 


\cutsubsectionup
\subsection{Virtual Cinematography}
\cutsubsectiondown
\stan{Finally, }existing virtual cinematography works \stanreplace{focus}{focused} on camera manipulation in simple virtual environments/video games~\cite{ChristiansonAHSWC96,He:1996,elson_lightweight_2007,Mindek:2015}\drliureplace{. They assumed the perception problem is bypassed (e.g., 3-D positions and poses of all entities are knowable).}{  and \stanreplace{do}{did} not deal with the perception difficulty problem. }\cite{foote2000flycam,sun2005region,chen2016learning,chen2015mimicking} relaxed the assumption and controlled virtual cameras within restricted static wide field-of-view video of a classroom, video conference, or basketball court, where objects of interest could be easily extracted. In contrast, our method handles raw 360$^\circ$ sports videos downloaded from YouTube\footnote{https://www.youtube.com/} in five domains (e.g., basketball, parkour, etc.). Recently, Su et al.~\cite{supano2vid} also proposed handling raw 360$^\circ$ videos download from YouTube. They referred to this problem as Pano2Vid --  automatic cinematography in 360$^\circ$ videos -- and proposed an offline method. In contrast, we propose an online human-like agent acting based on both present and previous observations. We argue that for handling streaming videos and other human-in-the-loop applications (e.g., foveated rendering\cite{Patney16}) a human-like online agent is necessary in order to provide more effective video-watching support. 
%
%


\cutsectionup
\section{Our Approach}\label{sec.Tech}
\cutsectiondown

We first define the 360 piloting problem in details (Sec.~\ref{sec.prob}). Then, we introduce our deep 360 pilot approach (Sec.~\ref{sec.dp}--Sec.~\ref{sec.full}). Finally, we describe the training procedure of our model (Sec.~\ref{sec.train}).

\begin{figure*}[t!]
\vspace{-3mm}
\begin{center}
\includegraphics[width=0.9\linewidth]{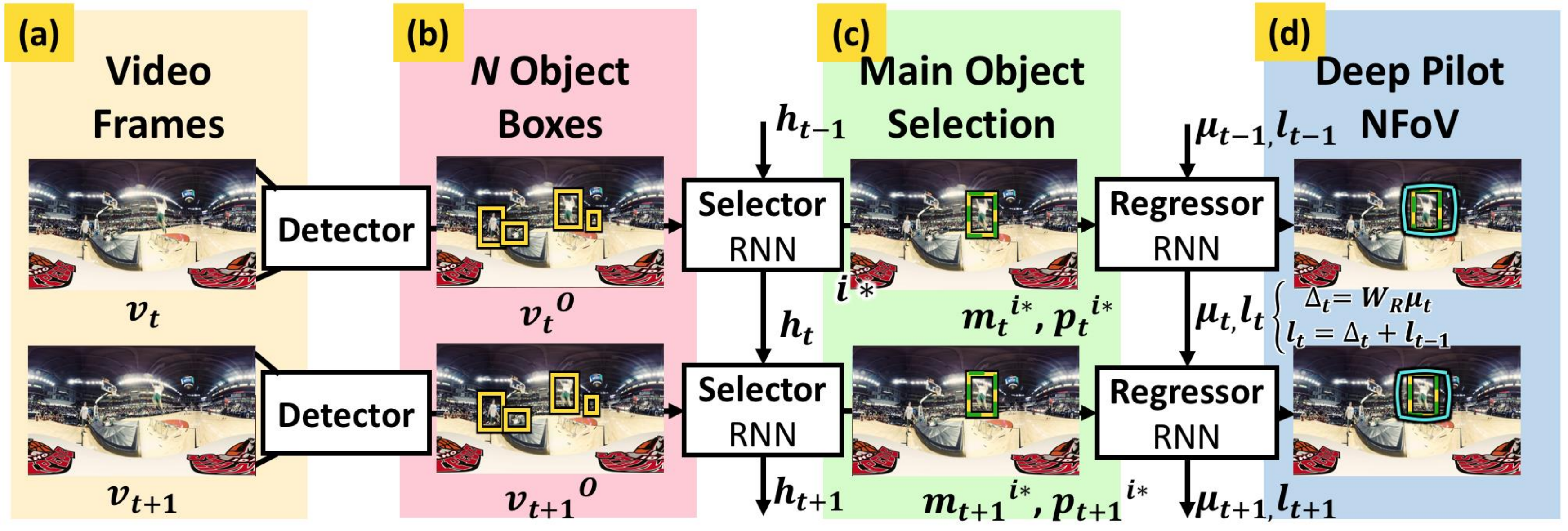}
\end{center}
\vspace{-3mm}
\caption{Visualization of our deep 360 pilot model. Panel (a) shows \drliureplace{the raw video frames in two consecutive frames}{two consecutive frames}. Panel (b) shows the top-$N$ confident \drliureplace{detected objects}{object bounding boxes} (yellow boxes) given by the detector. Panel (c) shows the selected main object (green dash box) given by the RNN-based Selector. Panel (d) shows the final NFoV centered at the viewing angle (cyan box) predicted by the RNN-based regressor.}\label{fig.model}
\vspace{-3mm}
\end{figure*}

\cutsubsectionup
\subsection{Definitions}\label{sec.prob}
\cutsubsectiondown

We formulate the 360 piloting task as the following online viewing angle selection task.

\noindent\textbf{Observation.}
At time $t$, the agent observes a new frame $v_t$, which is the $t$-th frame of the 360$^\circ$ video. The sequence of frames that the agent has observed up to this time is referred to as $\textbf{V}_t=\{v_1,...,v_t\}$.

\noindent\textbf{Goal.}
The goal of the agent is to select a viewing angle $l_t$ at time $t$ so that the sequence of viewing angles $\textbf{L}_t=\{l_1,...,l_t\}$ smoothly capture events of interest in the $360^\circ$ video. Note that $l_t=(\theta_t,\phi_t)$ is a point on the $360^\circ$ viewing sphere, parameterized by the azimuth angle $\theta_t\in [0^\circ,360^\circ]$ and elevation angle $\phi_t\in [-90^\circ,90^\circ]$

\noindent\textbf{Action.}
In order to achieve the goal, the agent takes the action of steering the viewing angle by $\Delta_t$ at time $t$. Given the previous viewing angle $l_{t-1}$ and current action $\Delta_t$, the current viewing angle $l_t$ is computed as follows,
\begin{eqnarray}
l_t = \Delta_t+l_{t-1}.\label{eq.al}
\end{eqnarray}


\noindent\textbf{Online policy.}
We assume that the agent takes an action $\Delta_t$ at frame $t$ according to an online policy $\pi$ as follows,
\begin{eqnarray}
\Delta_t = \pi(\textbf{V}_t,\textbf{L}_{t-1}),\label{eq.p}
\end{eqnarray}
where the online policy depends on both the current and previous observation $\textbf{V}_t$ and previous viewing angles $\textbf{L}_{t-1}$. 
This implies that the previous viewing angles affect the current action similar to what a human viewer acts when viewing a $360^\circ$ sports video.
Hence, the main task of 360 piloting is about  learning the online policy from data.

In the following, 
we discuss various design choices of our proposed deep 360 pilot where the online policy in Eq.~\ref{eq.p} is modeled as a deep neural network.

\cutsubsectionup
\subsection{Observing in Object Level}\label{sec.dp}
\cutsubsectiondown

Instead of extracting information from the whole 360$^\circ$ panoramic frame at each time instance, we propose to focus on foreground objects (Fig.~\ref{fig.model}(b)) for two reasons.
Firstly, in sports videos, foreground objects are typically the targets to be followed. Moreover, the relative size of foreground objects is small compared to the whole panoramic image. If processing is done at the frame level, information of object fine details would be diluted. Working with object-based observations help our method extract subtle appearance and motion cues to take an action. We define object-level observation $\textbf{V}^O_t$ as,
\begin{align}
&\textbf{V}^O_t=\{v^O_1,...,v^O_t\}\\
&\text{where $v^O_t$ is given by } v^O_t= \textrm{con}_V(O_t,P_t,M_t).\label{Eq.o}\\
&\text{and } O_t=\textrm{con}_H(\{o^i_t\}), P_t=\textrm{con}_H(\{p^i_t\}),\label{Eq.OandP}\\ 
&M_t=\textrm{con}_H(\{m^i_t\}).
\end{align}
Note that $\textrm{con}_H()$ and $\textrm{con}_V()$ denote horizontal and vertical concatenation of vectors, respectively. The vector $o^i_t\in R^{d}$ denotes the $i$-th object appearance feature\hureplace{and}{,} the vector $p_t^i\in R^2$ denotes the $i$-th object location (the same parameterization as $l_t$) on the view sphere at frame $t$ \hu{and the vector $m_t^i\in R^{k}$} \drliu{denotes the $i$-th object motion feature}.
If there are $N$ objects, \hureplace{the dimension of $O_t$ is $d\times N$, the dimension of $P_t$ is $2\times N$, and the dimension of $v^O_t$ is $(d+2+12)\times N$.}{the dimension of $O_t$, $P_t$, and $M_t$ are $d\times N$, $2\times N$, and $k\times N$, respectively. Then the dimension of concatenated object feature $v^O_t$ is $(d+2+k)\times N$.} \hu{Note that our agent is invariant to the order of objects. More explanation is shown in technical report~\cite{HuCVPRsupp17}}. In the online policy (Eq.~\ref{eq.p}), we replace $\textbf{V}_t$ with $\textbf{V}^O_t$ which consists of object appearance\drliu{, motion,} and location. 

\subsection{Focusing on the Main Object}\label{sec.mo}

We know that as watching a sports video a human agent gazes at the main object of interest.
Assuming the location of the main object of interest,  $p_t^{i*}$, is known, a naive policy for 360 piloting would be a policy that closely follows the main object and the action taken at each time instance is 
\vspace{-1mm}
\begin{eqnarray}
\hat{\Delta}_t=p_t^{i*}-l_{t-1}.\label{Eq.d}
\end{eqnarray}
\vspace{-1mm}
Since a machine agent does not know which object is the main one, we propose the following method to estimate the index $i*$ of the main object. We treat this task as a classification task and predict the probability $S_t(i)$ that the object $i$ is the main object as follows,
\vspace{-1mm}
\begin{eqnarray}
S_t=\pi(\textbf{V}^O_t)\in [0,1]^N,\label{eq.s}
\end{eqnarray}
\vspace{-1mm}
where $\sum_i S_t(i) = 1$.
Given $S_t$,
\vspace{-1mm}
\begin{eqnarray}
i*=\arg\max_i S_t(i).\label{eq.istar}
\end{eqnarray}
\vspace{-1mm}
In this case, the agent's task becomes {\it discretely} selecting one main object (Fig.~\ref{fig.model}(c)). We will need to handle this discrete selecting while introducing policy gradient~\cite{Williams1992}.

\sunmin{We note that the size of $\textbf{V}^O_t$ grows with the number of observed frames, which increase the computation cost. We propose to aggregate object previous information via a Recurrent Neural Network (RNN).}


\cutsubsectionup
\subsection{Aggregating Object Information}\label{sec.rnn}
\cutsubsectiondown

Our online policy is implemented as a selector network as shown in Fig.~\ref{fig.model}(b)). It consists of a RNN followed by a softmax layer. The RNN aggregates information from the current frame and past state to update its current state, while the softmax layer maps the current state of the RNN into a probability distribution via $W_s$. 
\begin{align}
&h_{t}=RNN_S(v^O_{t},h_{t-1}),\nonumber\\
&S_t =\textrm{softmax}(W_s h_t)\label{eq.h}
\end{align}

\cutsubsectionup
\subsection{\hureplace{Aggregating Object Motion Information}{Learning Smooth Transition}}
\cutsubsectiondown

\hureplace{So far the action is taken without considering the motion $m_t^{i*}$ of the main object.}{So far our model dose not take care of the smooth transition in viewing angle.} Hence, we propose to refine the action from the selector network, $\hat{\Delta}_t = p_t^{i*}-l_{t-1}$, with the motion feature, $m_t^{i*}$ (Fig.~\ref{fig.model}(d)), as follows,
\begin{eqnarray}
\mu_t &=& RNN_R(\textrm{con}_V(m_t^{i*},\hat{\Delta}_t),\mu_{t-1}).\nonumber\\
\Delta_t  &=& W_R \textrm{ } \mu_t,
\label{Eq.delta}
\end{eqnarray}
Here, we concatenate the motion feature and the proposed action from the selection network to form the input at time $t$ to the regressor network $RNN_R$. The $RNN_R$ then updates its state from $\mu_{t-1}$ to $\mu_{t}$. 
While $RNN_S$ focuses on main object selection, $RNN_R$ focuses on action refinement. The state of $RNN_R$ is then mapped to the final steering action vector $\Delta_t$ via $W_R$. The resulting viewing angle is then given by $l_t=\Delta_t+ l_{t-1}$.


\subsection{Our Final Model}\label{sec.full}
\cutsubsectiondown

As shown in Fig.~\ref{fig.model}, our model has three main blocks. The \textit{detector} block extracts object-based observation $v_t^o$ as described in Eq.~\ref{Eq.o}. The \textit{selector} block selects the main object index $i*$ following Eq.~\ref{eq.h} and Eq.~\ref{eq.istar}. The \textit{regressor} block \hureplace{regrees}{regresses} the viewing angle $l_t$ given main object location $p_t^{i*}$ and motion $m_t^{i*}$ following Eq.~\ref{Eq.d}, Eq.~\ref{Eq.delta}, and Eq.~\ref{eq.al}.

\cutsubsectionup
\subsection{Training}\label{sec.train}
\cutsubsectiondown

\drliureplace{In the following, we discuss the training of the proposed deep 360 pilot in details. We}{We} will first discuss the training of the regressor network and then discuss the training of the selector network. Finally, we show how to train these two networks jointly. Note that we use the viewing angle $l_t^{gt}$ at each time instance provided by human annotators as the ground truth.

\noindent\textbf{Regressor network.}
We train the regressor network by minimizing the Euclidean distance between the predicted viewing angle and the ground truth viewing angle at each time instance. For enforcing a smooth steering, we also regularize the training with a smoothness term, which penalizes a large rate of change in viewing angles between two consecutive frames. Let $v_t = l_{t} - l_{t-1}$ be the viewing angle velocities at time $t$. The loss function is then given by 
\vspace{-1mm}
\begin{equation}
\sum_{t=1}^{T} \|l_t-l_t^{gt}\|_2+\lambda \|v_t-v_{t-1}\|_2
\end{equation}
\vspace{-1mm}
where $\lambda$ is a hyper-parameter balancing the two terms \drliu{and $T$ is the number of frames in a video.}

\noindent\textbf{Selector network.}
\drliureplace{Note that}{As} the ground truth annotation for each frame is provided as the human viewing angle\drliureplace{. The}{, the} main object $i*^{gt}$ to be focused on at each frame is unknown. \drliureplace{As a result}{Therefore}, we resort to the approximated policy gradient technique proposed in~\cite{Williams1992} to train the selector network.
\\
Let $l(i)$ be a viewing angle associated with object $i$ that is computed by the regressor network. We define the reward of selecting object $i$ (steering the viewing angle to $l(i)$) to be $r(l(i))$ where the reward function $r$ is defined based on the overlapping ratio between the NFOV centering at $l_t^{i*}$ and the NFOV centering at $l(i)$. The details of the reward function design is shown in \hureplace{Fig. ???.}{technical report~\cite{HuCVPRsupp17}.} We then train the selector network by maximizing the expected reward
\begin{equation}
\mathcal{E}(\theta)=E_{i\sim S(i,\theta)}[r(l(i))],\label{eq.r}
\end{equation}
using the policy gradient
\begin{eqnarray}
\nabla_{\theta}\mathcal{E}(\theta)&=&\nabla_{\theta} E_{i\sim S(i,\theta)}[r(l(i))]\\
&=& E_{i\sim S(i,\theta)}[r(l(i)) \nabla_{\theta}\log S(i,\theta)],
\end{eqnarray}
where $\theta$ is the model parameter \drliureplace{for}{of} the selector network.

We further approximate $\nabla_{\theta}\mathcal{E}(\theta)$ using sampling as,
\vspace{-1mm}
\begin{eqnarray}
\nabla_{\theta}\mathcal{E}(\theta)\simeq\frac{1}{Q} \sum_{q=1}^{Q}r(l(i_q)) \nabla_{\theta}\log S(i_q,\theta),
\end{eqnarray}
\vspace{-1mm}
where $q$ is the index of sampled main object, $Q$ is the number of samples, and the approximated gradient is referred to as \drliu{the} policy gradient.
\drliureplace{
Since the policy gradient can be computed efficiently using back-propagation, we use policy gradient to maximize the expected reward.
}{}

\noindent\textbf{Joint training.}
Since the location of the object selected by the selector network is fed into the regressor network for computing the final viewing angle and the reward function for training the selector network is based on the regressor network's output, the two networks are trained jointly. Specifically, we joint update the trainable parameters in both networks similar to~\cite{mnih-attention-2014}, which hybrids the gradients from the reinforcement signal and supervised signal.

\cutsectionup
\section{\hureplace{Dataset}{Sports-360 Dataset}}\label{sec.Dataset}
\cutsectiondown

\begin{figure*}[t!]\cuttableup
\begin{center}
\includegraphics[width=1.00\linewidth]{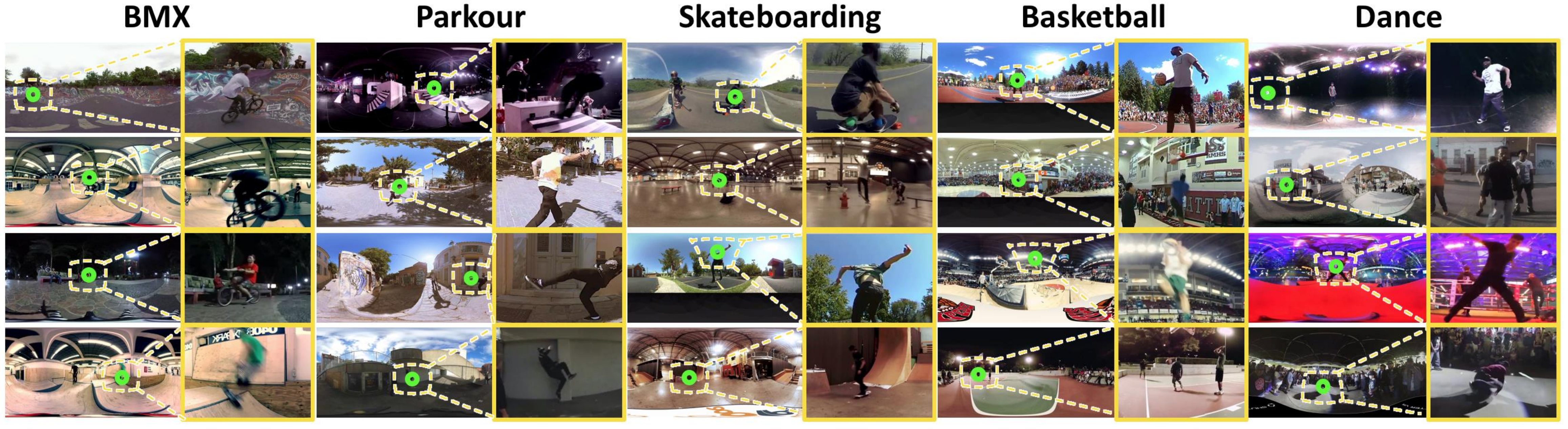}
\end{center}
\vspace{-3mm}
\caption{Our Sports-360 dataset. We show example pairs of panoramic \hureplace{frame and zoomed in }{and }NFoV images in five domains: BMX, parkour, skateboarding, basketball, and dance. In each example, a panoramic frame with ground truth viewing angle (green circle) is shown on the left. The zoomed in NFoV (yellow box) centered at the ground truth viewing angle is shown on the right. The NFoV illustrates the viewer’s perspective.}\label{fig.data}
\vspace{-4mm}
\end{figure*}

\begin{figure}[t!]
\begin{center}
\includegraphics[width=0.975\linewidth]{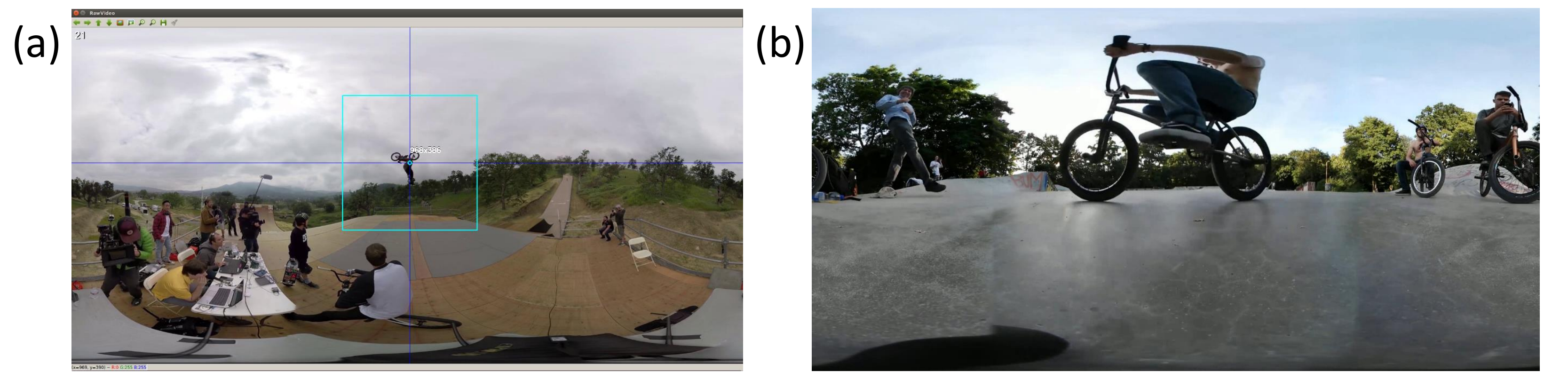}
\end{center}
\vspace{-3mm}
\caption{\footnotesize (a) Annotators mark main objects in 360$^\circ$ videos with a mouse. The blue cross helps annotators locate cursor position, and the cyan box indicates NFoV. \hu{Main reason to label in panorama is shown in the technical report~\cite{HuCVPRsupp17}.}
 (b) Example of bmx bike.}\label{fig.label}
\vspace{-4mm}
\end{figure}

We have collected a new dataset called Sports-360\hu{\footnote{Our dataset and code can be downloaded from \url{https://aliensunmin.github.io/project/360video}}}, which consists of 342 360$^{\circ}$ videos 
downloaded from YouTube in five sports domains: basketball, parkour, BMX, skateboarding, and dance (Fig.~\ref{fig.data}). 
Domains were selected according to the following criteria: (i) \stanreplace{we can retrieve a large number of 
relevant 360$^{\circ}$ videos on YouTube,}{high availability of such videos on YouTube,} (ii) the retrieved videos contain dynamic activities rather than static scenes, and (iii) 
\stanreplace{human can identify a clear object of interest in most frames.}{containing a clear human-identifiable object of interest in most of the video frames.}
The third criterion is required \hureplace{in order }{}to obtain unambiguous ground truth viewing angle in \hureplace{all of }{}our videos.

In each domain, we downloaded the top 200 videos sorted by relevance.
Then, we \stanreplace{filter out ones with}{removed videos that were either in} poor resolution 
or stitching quality.
Next, we \stanreplace{sample a continuous segment of each video with no scene transition since many of the 360$^{\circ}$ videos are edited and contain scene transition.}{sampled and extracted a continuous video clip from each video where a scene transition is absent (many 360$^{\circ}$ videos are edited and contain scene transitions).} 
\hu{Finally, we recruited 5 human annotators, and 3 were asked to "label the most salient object for VR viewers" in each frame in a set of video segments containing human-identifiable objects. Each video segment was annotated by 1 annotator in the panorama view (see Fig.~\ref{fig.label}a). The annotation results were verified and corrected by the other 2 annotators.}

\sunmin{We show example panoramic frames and NFoV images centered at ground truth viewing angles in Fig.~\ref{fig.data}.}
\stanreplace{Both the video segments and their corresponding ground truth viewing angles become our dataset.}{Our dataset includes both video segments and their annotated ground truth viewing angles.} The statistics of our dataset (i.e., number of videos and frames per domain) is shown in Table.~\ref{tab.IAexp}.
We split them by assigning 80\% of the videos for training, and 20\% for testing.

\begin{table}
\begin{center}
\small
\begin{tabular}[t]{|c||c|c|c|c|c|c|}
\hline
& SB & Park. & BMX & Dance & BB& Total\\
\hline
$\#$Video & 56 & 92 & 53 & 56 & 85 & 342\\
\hline
$\#$Frame & 59K & 27K & 16K & 56K & 22K & 180K\\
\hline
\end{tabular}
\normalsize
\end{center}
\vspace{-2mm}
\caption{\small Statistics of our Sports-360 dataset. SB, Park., BMX, and BB stand for skateboarding, parkour, bicycle motocross, and basketball, respectively. K stands for thousand.}\label{tab.IAexp}
\vspace{-4mm}
\end{table}


\cutsectionup
\section{Experiments}\label{sec.Exp}
\cutsectiondown

We evaluate deep 360 pilot on the Sports-360 dataset\drliureplace{ as well as conduct a user preference study.}{.} We show that our \hureplace{deep 360 pilot}{model} outperforms baselines by a large margin both quantitatively and qualitatively. \drliu{In addition, we also conduct a user preference study.}
In the following, we first define the evaluation metric. Then, we describe the implementation details and baseline methods\hureplace{ to be compared}{}.
Finally, we report the quantitative, qualitative, and human study results.




\cutsubsectionup
\subsection{Evaluation Metrics.}\label{sec.metric}
\cutsubsectiondown
To quantify our results, we report both Mean Overlap (MO) and Mean Velocity Difference (MVD). \textbf{MO} measures how much the NFoV centered at the predicted viewing angle overlaps (i.e., Intersection over Union (IoU)) with \drliu{that of} the ground truth one at each frame. A prediction is precise if the IoU \drliureplace{with the ground truth}{} is \drliureplace{high}{1}. \textbf{MVD} evaluates the curvature of the predicted viewing angle trajectories. \drliureplace{We define the viewing angle velocity as $\textrm{v}_t = l_{t}-l_{t-1}$ degrees per frame. The velocity difference is defined as $\|\textrm{v}_t-\textrm{v}_{t-1}\|_{2}$.}{It is given by the norm of the difference of viewing angle velocities in two consecutive frames given by $\|v_t-v_{t-1}\|_{2}$.}
Note that, in average, the trajectory is smoother if its \hureplace{mean velocity difference}{MVD} at each frame is low.

\cutsubsectionup
\subsection{Implementation Details}\label{sec.imp}
\cutsubsectiondown

\noindent\textbf{Detector.}
We use the Faster R-CNN~\cite{ren2015faster} model pre-trained on 2014 COCO detection dataset~\cite{lin2014microsoft} to generate about $400$ bounding boxes for each frame. Then, we apply the tracking-by-detection algorithm~\cite{Andriluka:CVPR08} to increase the recall of the object detection. Finally, we apply detection-by-tracking \cite{Andriluka:CVPR08} to select reliable detection linked into long tracklets.
Given these tracklets, we select top $N=16$ reliable boxes per frame as our object-based observation. \hu{Detailed sensitivity experiment results can be found in the technical report~\cite{HuCVPRsupp17}}. \hu{We found it is beneficial to use general object detectors. In the sport video domains studied, non-human objects such as skateboard, basketball, or bmx bike (Fig.~\ref{fig.label}b) provides strong cues for main objects. 
} For each object, we extract mean pooling of the $\textrm{Conv}5$ feature $\in R^{512}$ in the network of R-CNN as the appearance feature $o_t^i$, and Histogram of Optical Flow~\cite{Dalal2006} \hu{of boxes} with 12 orientation bins as the motion representation $m_t^i\in R^{12}$.

\noindent\textbf{Selector.}
\hureplace{We first compress the input $v^O_t\in R^{(d+2+12)\times N}$ into $R^{256}$ to reduce the complexity of the observation.
The hidden representation of $RNN_S$ is set to $256$.}{
The hidden representation of $RNN_S$ is set to $256$ and it processes 
input $v^O_t\in R^{(d+2+k)\times N}$ in sequences of 50 frames.}

\noindent\textbf{Regressor.}
The hidden representation of $RNN_R$ is set to $8$. We set $\lambda$ to $10$\drliureplace{without much manually tuning.}{.}

\noindent\textbf{Learning.}
We \drliureplace{implement}{optimize} our model \drliureplace{with}{using stochastic gradients with} batch size = $10$ \drliu{and} maximum epochs = $400$. \drliureplace{every $50$ epochs learning rate decay = $0.9$, and initial learning rate = $1e-5$.}{The learning rate is decayed by a factor of $0.9$ from the initial learning rate of $1e^{-5}$ every $50$ epochs.}

\cutsubsectionup
\subsection{Methods to be Compared}\label{sec.base}
\cutsubsectiondown
\stanreplace{We compare different variants of our method with state-of-the-art method and other baseline methods combining saliency detection with the object detector~\cite{ren2015faster}.}{We compared \drliu{the proposed deep} 360 pilot with a number of methods, including the state-of-the-art method AUTOCAM~\cite{supano2vid}, two baseline methods combining saliency detection with the object detector~\cite{ren2015faster},} and a variant of \drliu{deep} 360 pilot without a regressor. 

\noindent\textbf{AUTOCAM~\cite{supano2vid}:} Since their \hureplace{dataset}{model} is not publicly available, we use the ground truth viewing angles to generate NFoV videos from our dataset. These NFoV videos are used to discriminatively assign interestingness on a set of pre-defined viewing angles at each frame in a testing video.
Then, AUTOCAM uses dynamic programming to select optimal sequence of viewing angles. 
Finally, the sequence of viewing angles is smoothed in a post-processing step. Note that since AUTOCAM proposes multiple paths for each video, we use ground truth in testing data to select top ranked sequence of viewing angles as the system's final output. This creates a strong ``offline" baseline.

\noindent\textbf{RCNN+Motion:} We first extract detected boxes' optical flow. Then, we use a simple motion saliency proposed by~\cite{actiontubes}, median flow, and HoF~\cite{Dalal2006} as features to train a gradient boosting classifier to select the box that is most likely to contain the main object. Finally, we use center of the box selected sequentially by the classifier as predictions.

\noindent\textbf{RCNN+BMS:} We leverage the saliency detector proposed by Zhang et al.~\cite{zhang2016exploiting} to detect the most salient region in a frame.
With the knowledge of saliency map, we can \sunmin{extract the max saliency scores in each 
 box as a score}. Then we emit the most salient box center sequentially as our optimal viewing angle trajectories.

\begin{table*}[t!]\cuttableup
	\centering
    \small
	\begin{tabular}{|c|c|c|c|c|c|c|c|c|c|c|}
		\hline    
       \multicolumn{1}{|c|}{\multirow{2}{*}{Method}} & \multicolumn{2}{c|}{Skateboarding} & \multicolumn{2}{c|}{Parkour} & \multicolumn{2}{c|}{BMX} & \multicolumn{2}{c|}{Dance} & \multicolumn{2}{c|}{Basketball}\\ 
       \cline{2-11}
    	& MO & MVD & MO & MVD & MO & MVD & MO & MVD & MO & MVD \\\hline
       Ours \textbf{w/o Regressor.} & \textbf{0.71} & 6.03 & \textbf{0.74} & 4.72 & \textbf{0.71} & 10.73 & \textbf{0.79} & 4.32 & \textbf{0.67} &  8.62 \\\hline
       Ours & 0.68 & \textbf{3.06} & \textbf{0.74} & \textbf{4.41} & 0.69 & \textbf{8.36} & 0.76 & \textbf{2.45} & \hureplace{0.6}{0.66} & \textbf{6.50} \\\hline\hline
       AUTOCAM~\cite{supano2vid} & \textit{0.56} & \textit{0.25} & \textit{0.56} & \textit{0.71} & \textit{0.47} & \textit{0.55} & \textit{0.73} & \textit{0.15} & 0.51 & \textit{0.66} \\\hline
       \textbf{RCNN+BMS.} & 0.25 & 37.5 & 0.2 & 30.8 & 0.22 & 32.4 & 0.24 & 40.5 & 0.2 & 25.27\\\hline
       \textbf{RCNN+Motion.} & 0.56 & 34.8 & 0.47 & 26.2 & 0.42 & 25.2 & 0.72 & 31.4 & \textit{0.54} & 25.2\\\hline 
	\end{tabular}
    \vspace{1mm}
	\caption{Benchmark experiment results. \hu{Except ``AUTOCAM" achieving a very low MVD through an offline process, ``Ours w/o Regressor" achieves the best MO (the higher the better) and ``Ours" achieves the best MVD (the lower the better). Most importantly, ``Ours" strikes a good balance between MO and MVD.}}\label{tab.Fexp}
\end{table*}


\begin{table*}[t!]
\begin{center}
\small
\begin{tabular}[t]{lccccc}
\hline
 & Skateboarding & Parkour & BMX & Dance & Basketball \\
\hline
Comparison & win / loss & win / loss & win / loss & win / loss & win / loss \\
\hline
vs AUTOCAM & 34 / 2 & 35 / 1 & 31 / 5 & 34 / 2 & 36 / 0\\
vs Ours \textbf{w/o Regressor} & 28 / 8 & 29 / 7 & 26 / 10 & 31 / 5 & 34 / 2\\
\hline
vs human & 15 / 21 & 10 / 26 & 7 / 29 & 14 / 22 & 7 / 29 \\
\hline
\end{tabular}
\vspace{-3mm}
\end{center}
\caption{User study results. For all of the five sports domains, our method is significantly
preferred over AUTOCAM and \hureplace{Our+w/o+Regressor}{Our \textbf{w/o Regressor}}. 
Also, it is comparable to expert human in skateboarding and dance.}\label{table.m}
\vspace{-3mm}
\end{table*}



\noindent\textbf{Ours w/o Regressor:} We test the performance of our \stanreplace{our approach}{deep 360 pilot} \yenchen{without regressor. It emits box center of the selected main object as prediction at each frame.} 

\cutsubsectionup
\subsection{Benchmark Experiments}
\begin{figure}[t!]
\begin{center}
\includegraphics[width=0.975\linewidth, height=0.90\linewidth]{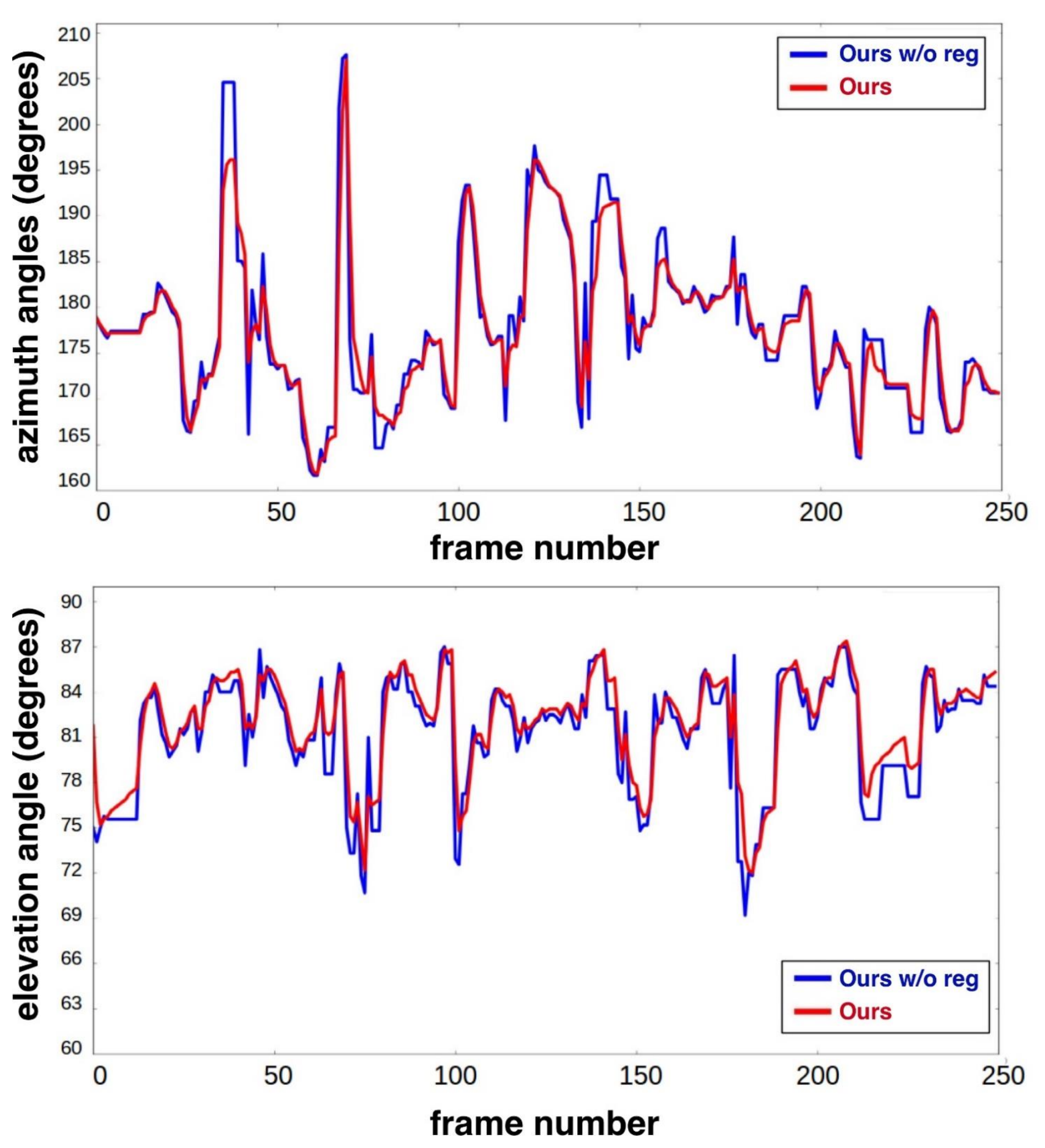}
\end{center}
\vspace{-3mm}
\caption{Comparison of “Ours” and “Ours w/o Regressor”. These two methods yields similar MO, while “Ours” predicts smoother viewing angles in both principal axes.}\label{fig.path}
\vspace{-3mm}
\end{figure}

\cutsubsectiondown
We compare our method with our variant and baseline methods in Table.~\ref{tab.Fexp}. In the following, we summarize our findings.
AUTOCAM achieves the best MO among three baseline methods in 4 out of 5 domains.
Our method significantly outperforms AUTOCAM in MO (at most $22\%$ gain in BMX and at least $3\%$ gain in Dance). 
Although AUTOCAM achieves significantly lower MVD compared to our method, we argue that its lower MO will critically affect its viewing quality, since the majority of our videos typically contain fast moving main objects. Since we do not know how to trade MVD over MO and vice versa, we resort to a user study to compare AUTOCAM with our method.
Our comparison with ours w/o regressor is the other way around. Both methods achieve similar MO while our method achieves lower MVD. These results show that with  regressor, the agent steers the viewing angle more smoothly.  \yenchen{Fig. \ref{fig.path} shows the trajectories of viewing angles predicted by both methods for a testing video. From this visual inspection, we verify that the smoothness term results in a less jittering trajectory.}

\begin{figure*}[t!]\cuttableup
\begin{center}
\includegraphics[width=0.99\linewidth]{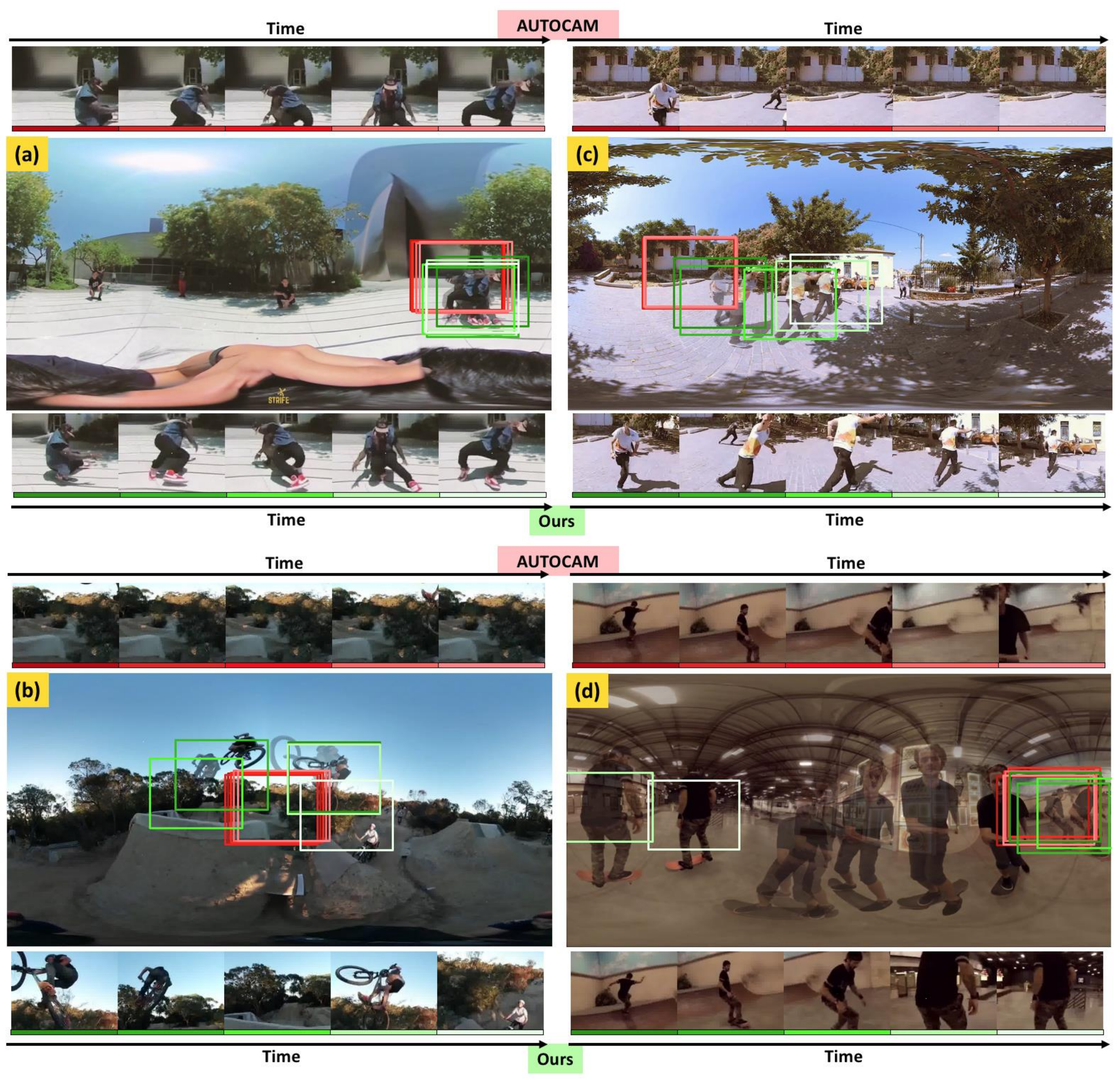}
\end{center}
\vspace{-3mm}
\caption{\chengreplace{Typical examples from four testing videos. For each video, the left panel shows the NFoV centered at three different viewing angles: AUTOCAM, Ours, and Ground Truth (GT). The right panel shows three different viewing angles overlaid on the panoramic image averaged across frames}{Typical examples from four domains: (a) dance, (b) BMX, (c) parkour, and (d) skateboarding. For each example, the middle panel shows a panoramic image with motaged foreground objects. The top and bottom panels show zoomed in NFoV centered at viewing angles generated by AUTOCAM and our method, respectively. We further overlaid the NFov from AUTOCAM and our method in red and green boxes, respectively, in the middle panoramic image. }}\label{fig.typ}
\vspace{-3mm}
\end{figure*}

\cutsubsectionup
\subsection{User Study}
\cutsubsectiondown
We conduct a user study mainly to compare our method with AUTOCAM and ours w/o regressor. 
The following is the experimental setting.
For each domain, we sample two videos where all three methods \chengreplace{achieves}{achieve} MO larger than $0.6$ and MVD smaller than $10$. This is to prevent users from comparing bad quality results, which makes identifying a better method difficult. For each video, we ask $18$ users to compare two methods. In each comparison, we show videos 
piloted by two methods with random order via a 360$^{\circ}$ 
video player. The number of times that our method wins or loses is shown in Table~\ref{table.m}.
Based on a two-tailed binomial test, our method is statistically superior \chengreplace{than}{to} AUTOCAM with p-value$<0.001$. This implies that users consider MO more important in this comparison.
Base on the same test, our method is statistically superior \chengreplace{than}{to} our w/o regressor with p-value$<0.05$. This implies that when MOs are similarly good, a small advantage of MVD results in a strong preference for our method. 
We also conduct a comparison between our method with the human labeled ground truth viewing angles.
Base on the same test, our method is indistinguishable to human on skateboarding with p-value$<0.405$ and on dance with p-value$<0.242$.

\cutsubsectionup
\subsection{Typical Examples}
\cutsubsectiondown

We compare our "deep 360 pilot" with AUTOCAM in Fig.~\ref{fig.typ}.
In the first example, both our method and AUTOCAM work well since the main object in dancing does not move globally. Hence, the ground truth viewing angle is not constantly moving.
In the next three examples, our method produces smooth trajectories while maintaining adequate view selection without any post-processing step. 
In contrast, AUTOCAM struggles on capturing fast-moving objects since Su et al.~\cite{supano2vid} constrains every glimpses' length up to 5 seconds. Moreover, the pre-defined $198$ views force many actions to be cut in half by the rendered NFoV. 
\hu{We further compare our method on a subset of publicly available videos from dataset of~\cite{supano2vid}. We get a 140\% performance boost in quantitative metrics of~\cite{supano2vid}.
Similar comparisons to other baseline methods and more results on dataset of~\cite{supano2vid} are shown in the technical report~\cite{HuCVPRsupp17}.}


\cutsectionup
\section{Conclusion}\label{sec.Con}
\cutsectiondown

We developed the first online agent for automatic 360\de video piloting. The agent was trained and evaluated using a newly composed Sport-360 dataset. \hu{We aimed at developing a domain-specific agent for the domain where the definition of a most salient object is clear (e.g., skateboarder).} The experiment results showed that our agent achieved much better performance as compared to the baseline methods including \cite{supano2vid}. \hu{However, our algorithm would suffer in the domains where our assumption is violated (containing equally salient objects or no objects at all).} In the future, we would like to reduce the amount of ground truth annotation needed for training our agent.

\cutsectionup
\section*{Acknowledgements}\label{sec.Ack} 
\vspace{-2mm}
We thank NOVATEK, MEDIATEK and NVIDIA for their support.
\vspace{-2mm}

{\small
\bibliographystyle{ieee}


\begin{thebibliography}{10}\itemsep=-1pt

\bibitem{AchantaHES09}
R.~Achanta, S.~S. Hemami, F.~J. Estrada, and S.~Süsstrunk.
\newblock Frequency-tuned salient region detection.
\newblock In {\em CVPR}, 2009.

\bibitem{Andriluka:CVPR08}
M.~Andriluka, S.~Roth, and B.~Schiele.
\newblock People-tracking-by-detection and people-detection-by-tracking.
\newblock In {\em CVPR}, 2008.

\bibitem{ba-attention-2015}
J.~Ba, V.~Mnih, and K.~Kavukcuoglu.
\newblock Multiple object recognition with visual attention.
\newblock In {\em ICLR'15}. 2015.

\bibitem{BourdevMalikICCV09}
L.~Bourdev and J.~Malik.
\newblock Poselets: Body part detectors trained using 3d human pose
  annotations.
\newblock In {\em ICCV}, 2009.

\bibitem{Bruce_2016_CVPR}
N.~D.~B. Bruce, C.~Catton, and S.~Janjic.
\newblock A deeper look at saliency: Feature contrast, semantics, and beyond.
\newblock In {\em CVPR}, June 2016.

\bibitem{chen2015mimicking}
J.~Chen and P.~Carr.
\newblock Mimicking human camera operators.
\newblock In {\em WACV}, pages 215--222. IEEE, 2015.

\bibitem{chen2016learning}
J.~Chen, H.~M. Le, P.~Carr, Y.~Yue, and J.~J. Little.
\newblock Learning online smooth predictors for realtime camera planning using
  recurrent decision trees.
\newblock In {\em CVPR}, 2016.

\bibitem{ChristiansonAHSWC96}
D.~B. Christianson, S.~E. Anderson, L.~wei He, D.~Salesin, D.~S. Weld, and
  M.~F. Cohen.
\newblock Declarative camera control for automatic cinematography.
\newblock In {\em AAAI}, 1996.

\bibitem{mlnet2016}
M.~Cornia, L.~Baraldi, G.~Serra, and R.~Cucchiara.
\newblock A deep multi-level network for saliency prediction.
\newblock In {\em ICPR}, 2016.

\bibitem{MSali09}
X.~Cui, Q.~Liu, and D.~Metaxas.
\newblock Temporal spectral residual: fast motion saliency detection.
\newblock In {\em ACM Multimedia}, 2009.

\bibitem{Dalal2006}
N.~Dalal, B.~Triggs, and C.~Schmid.
\newblock Human detection using oriented histograms of flow and appearance.
\newblock In {\em ECCV}, 2006.

\bibitem{elson_lightweight_2007}
D.~K. Elson and M.~O. Riedl.
\newblock A {Lightweight} {Intelligent} {Virtual} {Cinematography} {System} for
  {Machinima} {Production}.
\newblock In {\em AIIDE}, 2007.

\bibitem{GazePlus}
A.~Fathi, Y.~Li, and J.~M. Rehg.
\newblock Learning to recognize daily actions using gaze.
\newblock In {\em ECCV}, 2012.

\bibitem{foote2000flycam}
J.~Foote and D.~Kimber.
\newblock Flycam: Practical panoramic video and automatic camera control.
\newblock In {\em ICME}, 2000.

\bibitem{actiontubes}
G.~Gkioxari and J.~Malik.
\newblock Finding action tubes.
\newblock 2015.

\bibitem{GofermanTPAMI}
S.~Goferman, L.~Zelnik-Manor, and A.~Tal.
\newblock Context-aware saliency detection.
\newblock {\em TPAMI}, 34(10):1915--1926, 2012.

\bibitem{StoryB06}
D.~Goldman, B.~Curless, D.~Salesin, and S.~Seitz.
\newblock Schematic storyboarding for video visualization and editing.
\newblock In {\em SIGGRAPH}, 2006.

\bibitem{NIPS2014_5413}
B.~Gong, W.-L. Chao, K.~Grauman, and F.~Sha.
\newblock Diverse sequential subset selection for supervised video
  summarization.
\newblock In {\em NIPS}, 2014.

\bibitem{LiuCVPR00}
Y.~Gong and X.~Liu.
\newblock Video summarization using singular value decomposition.
\newblock In {\em CVPR}, 2000.

\bibitem{STSali08}
C.~Guo, Q.~Ma, and L.~Zhang.
\newblock Spatio-temporal saliency detection using phase spectrum of quaternion
  fourier transform.
\newblock In {\em CVPR}, 2008.

\bibitem{HarelKP06}
J.~Harel, C.~Koch, and P.~Perona.
\newblock Graph-based visual saliency.
\newblock In {\em NIPS}, 2006.

\bibitem{He:1996}
L.-w. He, M.~F. Cohen, and D.~H. Salesin.
\newblock The virtual cinematographer: A paradigm for automatic real-time
  camera control and directing.
\newblock In {\em ACM CGI}, 1996.

\bibitem{HuCVPRsupp17}
H.-N. Hu, Y.-C. Lin, M.-Y. Liu, H.-T. Cheng, Y.-J. Chang, and M.~Sun.
\newblock Technical report of deep 360 pilot.
\newblock 2017.
\newblock \url{https://aliensunmin.github.io/project/360video}.

\bibitem{Jetley_2016_CVPR}
S.~Jetley, N.~Murray, and E.~Vig.
\newblock End-to-end saliency mapping via probability distribution prediction.
\newblock In {\em CVPR}, 2016.

\bibitem{Cliplets12}
N.~Joshi, S.~Metha, S.~Drucker, E.~Stollnitz, H.~Hoppe, M.~Uyttendaele, and
  M.~F. Cohen.
\newblock Cliplets: Juxtaposing still and dynamic imagery.
\newblock In {\em UIST}, 2012.

\bibitem{Judd09}
T.~Judd, K.~Ehinger, F.~Durand, and A.~Torralba.
\newblock Learning to predict where humans look.
\newblock In {\em ICCV}, 2009.

\bibitem{CVPR13_Khosla}
A.~Khosla, R.~Hamid, C.-J. Lin, and N.~Sundaresan.
\newblock Large-scale video summarization using web-image priors.
\newblock In {\em CVPR}, 2013.

\bibitem{Kopf:FHV}
J.~Kopf, M.~F. Cohen, and R.~Szeliski.
\newblock First-person hyper-lapse videos.
\newblock {\em ACM Trans. Graph.}, 33(4), July 2014.

\bibitem{lee2015low}
T.~Lee, M.~Hwangbo, T.~Alan, O.~Tickoo, and R.~Iyer.
\newblock Low-complexity hog for efficient video saliency.
\newblock In {\em ICIP}, pages 3749--3752. IEEE, 2015.

\bibitem{EgoS12}
Y.~J. Lee, J.~Ghosh, and K.~Grauman.
\newblock Discovering important people and objects for egocentric video
  summarization.
\newblock In {\em CVPR}, 2012.

\bibitem{lin2014microsoft}
T.-Y. Lin, M.~Maire, S.~Belongie, J.~Hays, P.~Perona, D.~Ramanan,
  P.~Doll{\'a}r, and C.~L. Zitnick.
\newblock Microsoft coco: Common objects in context.
\newblock In {\em ECCV}, 2014.

\bibitem{LinCHI17}
Y.-C. Lin, Y.-J. Chang, H.-N. Hu, H.-T. Cheng, C.-W. Huang, and M.~Sun.
\newblock Tell me where to look: Investigating ways for assisting focus in
  360° video.
\newblock In {\em CHI}, 2017.

\bibitem{LiuTPAMI}
D.~Liu, G.~Hua, and T.~Chen.
\newblock A hierarchical visual model for video object summarization.
\newblock {\em TPAMI}, 32(12):2178--2190, 2010.

\bibitem{Liu_2016_CVPR}
N.~Liu and J.~Han.
\newblock Dhsnet: Deep hierarchical saliency network for salient object
  detection.
\newblock In {\em CVPR}, 2016.

\bibitem{liu2011learning}
T.~Liu, Z.~Yuan, J.~Sun, J.~Wang, N.~Zheng, X.~Tang, and H.-Y. Shum.
\newblock Learning to detect a salient object.
\newblock {\em TPAMI}, 33(2):353--367, 2011.

\bibitem{EgoS13}
Z.~Lu and K.~Grauman.
\newblock Story-driven summarization for egocentric video.
\newblock In {\em CVPR}, 2013.

\bibitem{MahadevanTPAMI}
V.~Mahadevan and N.~Vasconcelos.
\newblock Spatiotemporal saliency in dynamic scenes.
\newblock {\em TPAMI}, 32(1):171--177, 2010.

\bibitem{mps-rlcvpr16}
S.~Mathe, A.~Pirinen, and C.~Sminchisescu.
\newblock {Reinforcement Learning for Visual Object Detection}.
\newblock In {\em CVPR}, June 2016.

\bibitem{MatheSminchisescuPAMI2015}
S.~Mathe and C.~Sminchisescu.
\newblock Actions in the eye: Dynamic gaze datasets and learnt saliency models
  for visual recognition.
\newblock {\em TPAMI}, 37, 2015.

\bibitem{Mindek:2015}
P.~Mindek, L.~\v{C}mol\'{\i}k, I.~Viola, E.~Gr\"{o}ller, and S.~Bruckner.
\newblock Automatized summarization of multiplayer games.
\newblock In {\em ACM CCG}, 2015.

\bibitem{DIEM}
P.~Mital, T.~Smith, R.~Hill, and J.~Henderson.
\newblock Clustering of gaze during dynamic scene viewing is predicted by
  motion.
\newblock {\em Cognitive Computation}, 3(1):5--24, 2011.

\bibitem{mnih-attention-2014}
V.~Mnih, N.~Heess, A.~Graves, and k.~kavukcuoglu.
\newblock Recurrent models of visual attention.
\newblock In Z.~Ghahramani, M.~Welling, C.~Cortes, N.~D. Lawrence, and K.~Q.
  Weinberger, editors, {\em NIPS}. 2014.

\bibitem{ZhanCSVT05}
C.~Ngo, Y.~Ma, and H.~Zhan.
\newblock Video summarization and scene detection by graph modeling.
\newblock In {\em CSVT}, 2005.

\bibitem{pan2016shallow}
J.~Pan, K.~McGuinness, E.~Sayrol, N.~O'Connor, and X.~Giro-i Nieto.
\newblock Shallow and deep convolutional networks for saliency prediction.
\newblock In {\em CVPR}, 2016.

\bibitem{Patney16}
A.~Patney, J.~Kim, M.~Salvi, A.~Kaplanyan, C.~Wyman, N.~Benty, A.~Lefohn, and
  D.~Luebke.
\newblock Perceptually-based foveated virtual reality.
\newblock In {\em SIGGRAPH}, pages 17:1--17:2, 2016.

\bibitem{perazzi2012saliency}
F.~Perazzi, P.~Kr{\"a}henb{\"u}hl, Y.~Pritch, and A.~Hornung.
\newblock Saliency filters: Contrast based filtering for salient region
  detection.
\newblock In {\em CVPR}, 2012.

\bibitem{potapov2014category}
D.~Potapov, M.~Douze, Z.~Harchaoui, and C.~Schmid.
\newblock Category-specific video summarization.
\newblock In {\em ECCV}, 2014.

\bibitem{WebSyn07}
Y.~Pritch, A.~Rav-Acha, A.~Gutman, and S.~Peleg.
\newblock Webcam synopsis: Peeking around the world.
\newblock In {\em ICCV}, 2007.

\bibitem{L2S06}
A.~Rav-Acha, Y.~Pritch, and S.~Peleg.
\newblock Making a long video short.
\newblock In {\em CVPR}, 2006.

\bibitem{ren2015faster}
S.~Ren, K.~He, R.~Girshick, and J.~Sun.
\newblock Faster r-cnn: Towards real-time object detection with region proposal
  networks.
\newblock In {\em Advances in neural information processing systems}, pages
  91--99, 2015.

\bibitem{rudoy2013learning}
D.~Rudoy, D.~B. Goldman, E.~Shechtman, and L.~Zelnik-Manor.
\newblock Learning video saliency from human gaze using candidate selection.
\newblock In {\em CVPR}, pages 1147--1154, 2013.

\bibitem{SeoJOV}
H.~Seo and P.~Milanfar.
\newblock Static and space-time visual saliency detection by self-resemblance.
\newblock {\em Journal of Vision}, 2009.

\bibitem{supano2vid}
Y.-C. Su, D.~Jayaraman, and K.~Grauman.
\newblock Pano2vid: Automatic cinematography for watching 360◦ videos.
\newblock In {\em ACCV}, 2016.

\bibitem{sun2014ranking}
M.~Sun, A.~Farhadi, and S.~Seitz.
\newblock Ranking domain-specific highlights by analyzing edited videos.
\newblock In {\em ECCV}, 2014.

\bibitem{MSunMontage}
M.~Sun, A.~Farhadi, B.~Taskar, and S.~Seitz.
\newblock Summarizing unconstrained videos using salient montages.
\newblock In {\em ECCV}, 2014.

\bibitem{sun2005region}
X.~Sun, J.~Foote, D.~Kimber, and B.~Manjunath.
\newblock Region of interest extraction and virtual camera control based on
  panoramic video capturing.
\newblock {\em TMM}, 7(5):981--990, 2005.

\bibitem{TangW16}
Y.~Tang and X.~Wu.
\newblock Saliency detection via combining region-level and pixel-level
  predictions with cnns.
\newblock In {\em ECCV}, 2016.

\bibitem{Truong:2007}
B.~T. Truong and S.~Venkatesh.
\newblock Video abstraction: A systematic review and classification.
\newblock {\em TMCCA}, 3(1), Feb. 2007.

\bibitem{wang2016learning}
J.~Wang, A.~Borji, C.-C.~J. Kuo, and L.~Itti.
\newblock Learning a combined model of visual saliency for fixation prediction.
\newblock {\em TIP}, 25(4):1566--1579, 2016.

\bibitem{WangWLZR16}
L.~Wang, L.~Wang, H.~Lu, P.~Zhang, and X.~Ruan.
\newblock Saliency detection with recurrent fully convolutional networks.
\newblock In {\em ECCV}, 2016.

\bibitem{Wang_2016_CVPR}
Q.~Wang, W.~Zheng, and R.~Piramuthu.
\newblock Grab: Visual saliency via novel graph model and background priors.
\newblock In {\em CVPR}, June 2016.

\bibitem{Williams1992}
R.~J. Williams.
\newblock Simple statistical gradient-following algorithms for connectionist
  reinforcement learning.
\newblock {\em Machine Learning}, 8(3):229--256, 1992.

\bibitem{Yao_2016_CVPR}
T.~Yao, T.~Mei, and Y.~Rui.
\newblock Highlight detection with pairwise deep ranking for first-person video
  summarization.
\newblock In {\em CVPR}, 2016.

\bibitem{zhang2016exploiting}
J.~Zhang and S.~Sclaroff.
\newblock Exploiting surroundedness for saliency detection: a boolean map
  approach.
\newblock {\em TPAMI}, 38(5):889--902, 2016.

\bibitem{ZhangCSG16}
K.~Zhang, W.~Chao, F.~Sha, and K.~Grauman.
\newblock Video summarization with long short-term memory.
\newblock In {\em ECCV}, 2016.

\bibitem{Zhang_2016_CVPR}
K.~Zhang, W.-L. Chao, F.~Sha, and K.~Grauman.
\newblock Summary transfer: Exemplar-based subset selection for video
  summarization.
\newblock In {\em CVPR}, June 2016.

\bibitem{BinLiveLight}
B.~Zhao and E.~Xing.
\newblock Quasi real-time summarization for consumer videos.
\newblock In {\em CVPR}, June 2014.

\end{thebibliography}
}

\clearpage
\appendix
\section{Reward Function}
Let $l_t(i)$ be the viewing angle associated with object $i$ that is computed by the regressor network, and $l_t^{gt}$ be the ground truth viewing angle at frame $t$. We define the reward function $r$ as follow,
\begin{equation}\small
    r(l_t(i), l_t^{gt})=
    \begin{cases}
      1-\frac{\|l_t(i)-l_t^{gt}\|_2}{\eta}, & \text{if}\ \|l_t(i)-l_t^{gt}\|_2<=\eta \\
      -1, & \text{otherwise}
    \end{cases}
\end{equation}

where $\eta$ equals the distance from the center of a viewing angle to the corner of its corresponding NFoV, i.e., $\sqrt[]{32.75^2 + 24.56^2}=40.9$ if we define NFOV as spanning a horizontal angle of 65.5$^\circ$ with a $4:3$ aspect ratio.
When $l_t==l_t^{gt}$, the reward is 1, which is the maximum reward. When $\|l_t(i)-l_t^{gt}\|_2>\eta$, i.e., the predicted viewing angle is not covered by ground truth viewing angle's NFoV, the reward is -1.

\section{Sensitivity Analysis}

In order to see if the number of candidate objects \chengreplace{effects}{affects} the performance of our system significantly, we conduct a sensitivity experiment on the number of candidate objects $N$. We evaluate our deep 360 pilot with $N=\{8,16,32\}$ in each domain. Experiment results in Table.~\ref{table.Senexp} suggests that deep 360 pilot is not sensitive to the number of candidate objects $N$. Also, for all the three values of $N$, deep 360 pilot still outperforms other baselines.

\vspace{-3mm}
\begin{table}[b!]
\begin{center}
\footnotesize
\begin{tabular}[t]{lccc}
\hline
&  & AUTOCAM & Ours (20 videos)\\
\hline
\multirow{2}{*}{
Similarity}
& Trajectory & 0.304 & \textbf{0.426} \\
& Frame & 0.581 & \textbf{0.764} \\
\hline
\multirow{2}{*}{ Overlap}
& Trajectory & 0.255 & \textbf{0.355} \\
& Frame & 0.389 & \textbf{0.560} \\
\hline
\end{tabular}
\end{center}
\vspace{-3mm}
\caption{\footnotesize Performance on ~\cite{supano2vid}.}
\vspace{-5mm}
\label{table.m}
\end{table}

\section{Typical Examples}

\begin{figure*}[t!]
\begin{center}
\includegraphics[width=0.975\linewidth]{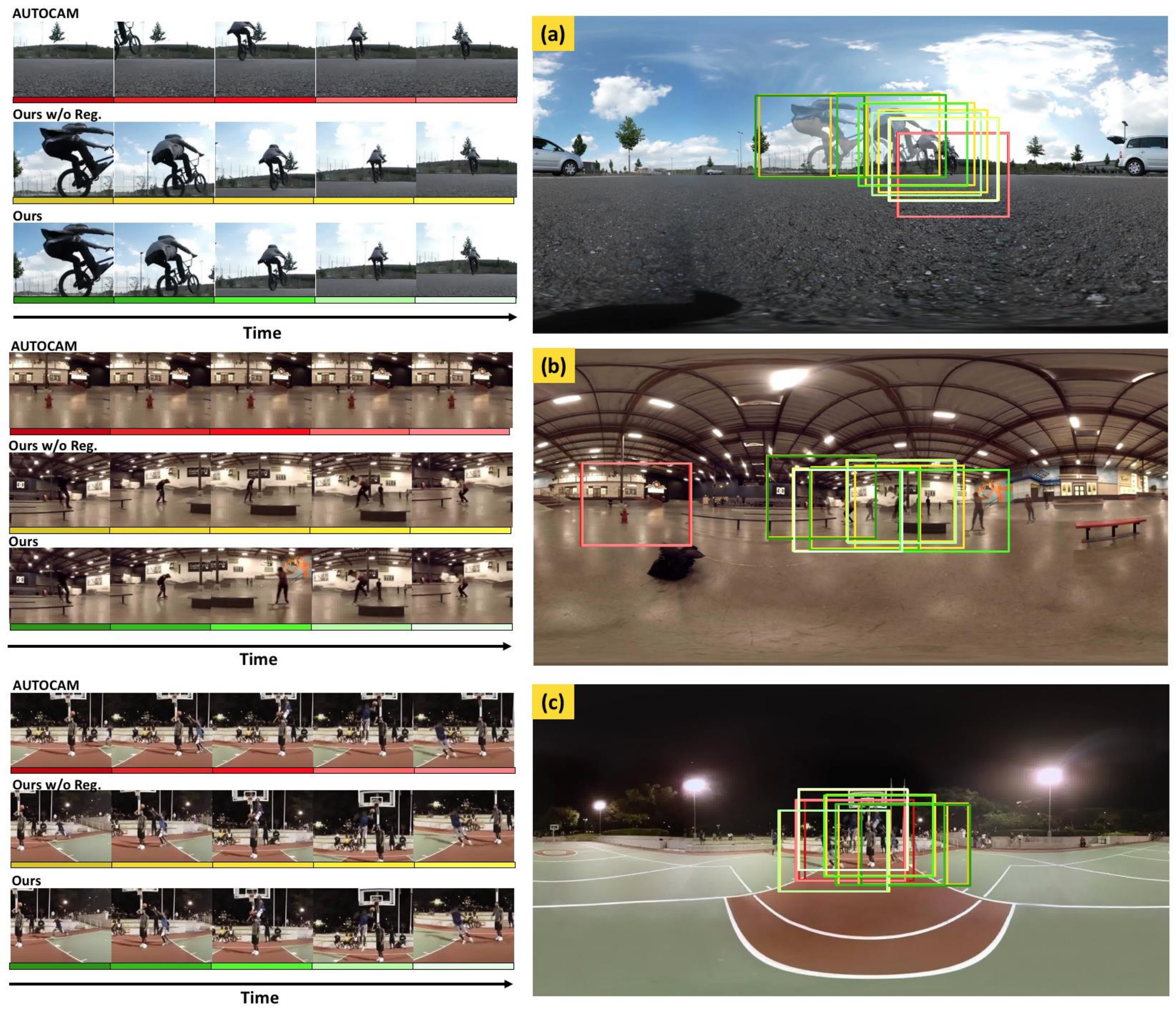}
\end{center}
\caption{Typical examples of three methods: AUTOCAM, Our method, and Our method without Regressor, from three domains: (a) BMX, (b) skateboarding, and (c) basketball. For each example, the right panel shows a panoramic image with montaged foreground objects. The left panel shows zoomed in NFoV centered at viewing angles generated by each \chengreplace{methods}{method}, respectively. We further overlaid the NFoV from AUTOCAM, Our method, and Ours without Regressor in red, green and yellow boxes, respectively, in the left panoramic image.}\label{fig.ex_1}
\end{figure*}

\begin{figure*}[t!]
\begin{center}
\includegraphics[width=0.975\linewidth]{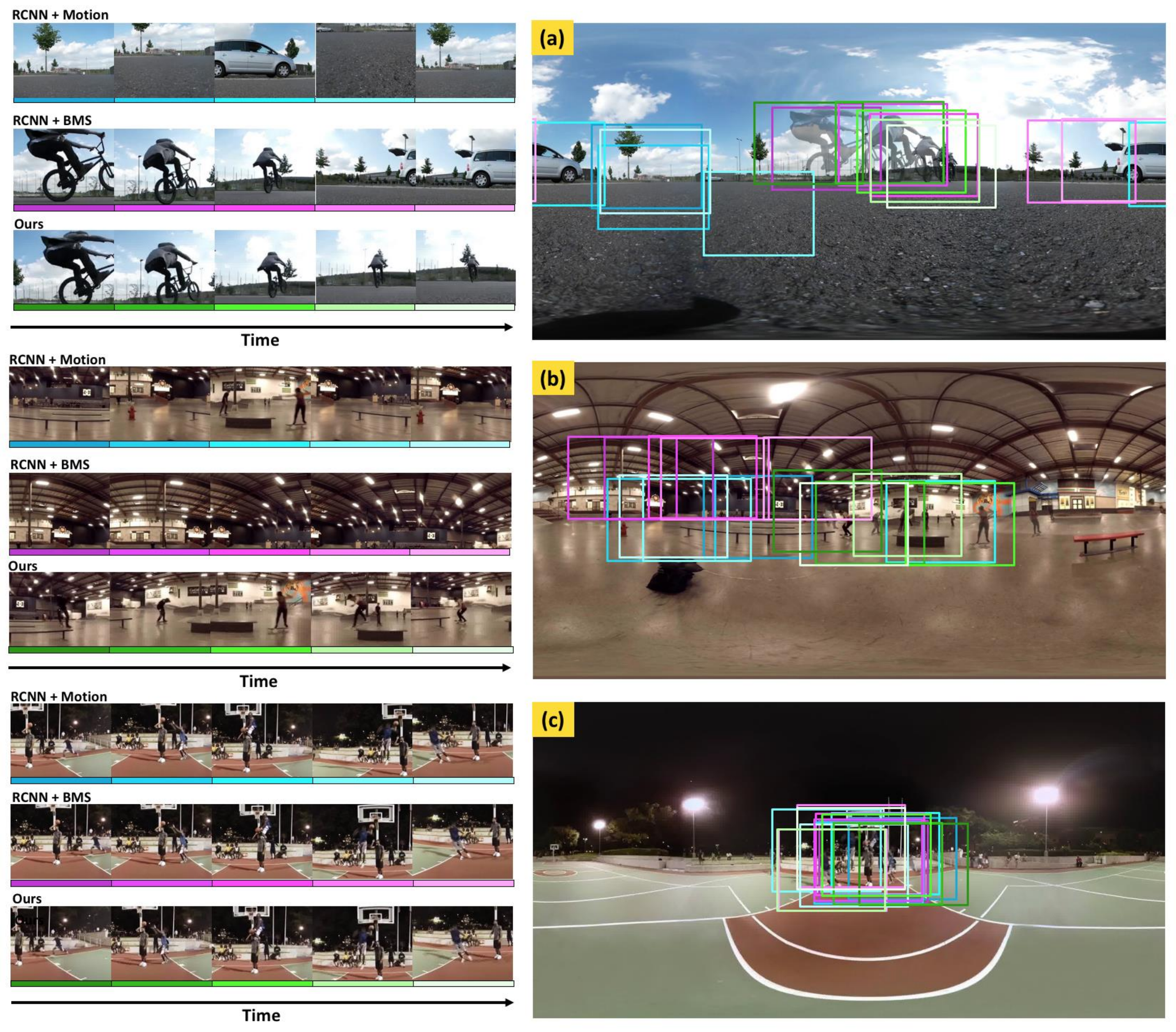}
\end{center}
\caption{Typical examples of three methods: RCNN + Motion, RCNN + BMS, and Our method, from three domains: (a) BMX, (b) skateboarding, and (c) basketball. Here we illustrate different results by the same way as in Fig.~\ref{fig.ex_1}, but overlaid the NFoV from RCNN + Motion, RCNN + BMS, and Our method in cyan, pink and green boxes. }\label{fig.ex_2}
\end{figure*}

We compare our "deep 360 pilot" method with several baselines: AUTOCAM, Our without Regressor in Fig.~\ref{fig.ex_1}, and RCNN + Motion, RCNN + BMS in Fig.~\ref{fig.ex_2}. Each method will generate a series of NFoV predictions on 3 videos: (a) a BMX video with a fast moving foreground object, (b) a skateboarding video with 2 main skateboarders, and (c) video of basketball players with relatively small movement. In Fig.~\ref{fig.ex_1} we can see that AUTOCAM almost stay at the same position, which is hard to follow the quick moving object in video (a) and (b), but our methods successfully capture the main foreground objects in each frame. In Fig.~\ref{fig.ex_2}, RCNN with either BMS saliency method or optical flow based method fail to stay focus on the main foreground objects in video (a) and (b), but our method is significantly outperforming these baselines. In the example of video (c) in both Figures, all predictions of different methods seem to be similar because of the smaller movement. However, our method still \chengreplace{capture}{captures} the main basketball player more precisely, where the player running from right to left to finish the slam dunk. 

\section{Human Evaluation Videos}
We upload a demo video which contains 3 examples selected from videos used for our human evaluation study, each of them comes from different domains. In each example, we demonstrate 4 methods, human label, AUTOCAM, ours, ours w/o Regressor, concurrently to make a clear comparison. This video can be found from \url{https://aliensunmin.github.io/project/360video}


\section{Reviewers' Comments}
We address critical comments from the reviewers below.

\subsection{Why not annotate in the Natural Field of View (NFoV)?} 
\hu{
We found that annotating directly in the NFoV is very inefficient because the annotators have to watch a video many times with different NFoVs. This involves a number of back-and-forth operations and makes the annotation process extremely tedious. In contrast, it is more efficient to compare two NFoV trajectories. Hence, we conducted a user study in NFoV, which matched the setting of our targeted use case.
}
\subsection{Apply our model on the dataset in AUTOCAM~\cite{supano2vid}.} 
\hu{
We cannot train our domain-specific agents using the dataset in~\cite{supano2vid} because the training videos are given in the NFOV format instead of the 360$^\circ$ format. Hence, we applied our model (trained for skateboarding) to all the 20 testing videos (downloaded on Jan. 30, 2017) provided with ground truths in the project page of~\cite{supano2vid}. Note that all the 20 testing videos are given in the 360$^\circ$ format but are in different domains (hiking, mountain climbing, parade, and soccer). We reported the results using metrics adopted by~\cite{supano2vid} in Table.~\ref{table.m}. Typical and failure example \hureplace{at: \url{https://youtu.be/u8nOEcFRcmI}}{ videos are available at \url{https://aliensunmin.github.io/project/360video}}. We found that our method achieved \hureplace{better performance. }{a 140\% performance boost of~\cite{supano2vid} in both similarity and overlap trajectory metric.}
}

\subsection{How order consistency are handled when objects disappear/reappear?}
\hu{
For each frame, the feature vectors of the objects are concatenated as a vector based on the order of their scores (See Eq.\hureplace{~\ref{Eq.OandP}}{5} in \chengreplace{}{the} main paper). When an object disappears or reappears, the concatenated vector does change. However, we empirically found that RNN seems to embed different vectors of similar scenes into similar points in the embedded space and did not suffer from this problem.
}

\begin{table*}[	t!]
	\centering
	\begin{adjustbox}{width=1.0\textwidth}
   \small
	\begin{tabular}{|c|c|c|c|c|c|c|c|c|c|c|}
		\hline    
      \multicolumn{1}{|c|}{\multirow{2}{*}{Our Method}} & \multicolumn{2}{c|}{Skateboarding} & \multicolumn{2}{c|}{Parkour} & \multicolumn{2}{c|}{BMX} & \multicolumn{2}{c|}{Dance} & \multicolumn{2}{c|}{Basketball}\\ 
      \cline{2-11}
   	& MO & MVD & MO & MVD & MO & MVD & MO & MVD & MO & MVD \\\hline
      N=8 & 0.68 & 2.99 & 0.69 & 3.71 & 0.65 & 8.58 & 0.74 & 2.53 & 0.67 &  5.36 \\\hline
      N=16 & \sunmin{0.68} & 3.06 & 0.74 & 4.41 & 0.69 & 8.36 & 0.76 & 2.45 & 0.66 & 6.50 \\\hline
      N=32 & 0.68 & 3.22 & 0.65 & 3.28 & 0.70 & 7.94 & 0.73 & 2.48 & 0.69 & 5.04 \\\hline
	\end{tabular}
   \end{adjustbox}
   \vspace{1mm}
	\caption{Sensitivity analysis of number of candidate objects $N$ on all five domains.}\label{table.Senexp}
\end{table*}

\end{document}